\documentclass{article}

\usepackage{multirow}
\usepackage{setspace}
\usepackage{amsmath}
\usepackage{amssymb}
\usepackage{bm}
\usepackage{placeins}
\usepackage{hyperref}
\usepackage{algorithm}
\usepackage{algorithmic}
\usepackage{lipsum}  
\usepackage[dvipsnames]{xcolor}
\usepackage{marvosym}
\usepackage{latexsym}
\usepackage{pifont}
\usepackage{diagbox}
\usepackage{url}            
\usepackage{amsfonts}       
\usepackage{nicefrac}       
\usepackage{wrapfig}
\usepackage{wasysym}
\usepackage{transparent}

\usepackage{microtype}
\usepackage{graphicx}
\usepackage{subfigure}
\usepackage{booktabs} 

\usepackage{hyperref}

\usepackage[accepted]{icml2021}

\icmltitlerunning{LogME: Practical Assessment of Pre-trained Models for Transfer Learning}

\begin{document}

\twocolumn[
\icmltitle{LogME: Practical Assessment of Pre-trained Models for Transfer Learning}

\icmlsetsymbol{equal}{*}

\begin{icmlauthorlist}
\icmlauthor{Kaichao You$^*$}{thss}
\icmlauthor{Yong Liu$^*$}{thss}
\icmlauthor{Jianmin Wang}{thss}
\icmlauthor{Mingsheng Long}{thss}
\end{icmlauthorlist}

\icmlaffiliation{thss}{School of Software, BNRist, Tsinghua University, Beijing 100084, China.

Kaichao You $<$youkaichao@gmail.com$>$}

\icmlcorrespondingauthor{Mingsheng Long}{mingsheng@tsinghua.edu.cn}

\icmlkeywords{Transfer Learning, Model Assessment, Maximum Evidence}

\vskip 0.3in
]

\printAffiliationsAndNotice{\icmlEqualContribution} 

\begin{abstract}
  This paper studies task adaptive pre-trained model selection, an underexplored problem of assessing pre-trained models for the target task and select best ones from the model zoo \emph{without fine-tuning}. A few pilot works addressed the problem in transferring supervised pre-trained models to classification tasks, but they cannot handle emerging unsupervised pre-trained models or regression tasks. In pursuit of a practical assessment method, we propose to estimate the maximum value of label evidence given features extracted by pre-trained models. Unlike the maximum likelihood, the maximum evidence is \emph{immune to over-fitting}, while its expensive computation can be dramatically reduced by our carefully designed algorithm. The Logarithm of Maximum Evidence (LogME) can be used to assess pre-trained models for transfer learning: a pre-trained model with a high LogME value is likely to have good transfer performance. LogME is \emph{fast, accurate, and general}, characterizing itself as the first practical method for assessing pre-trained models. Compared with brute-force fine-tuning, LogME brings at most $3000\times$ speedup in wall-clock time and requires only $1\%$ memory footprint. It outperforms prior methods by a large margin in their setting and is applicable to new settings. It is general enough for diverse pre-trained models (supervised pre-trained and unsupervised pre-trained), downstream tasks (classification and regression), and modalities (vision and language). Code is available at this repository: \href{https://github.com/thuml/LogME}{https://github.com/thuml/LogME}.
\end{abstract}

\section{Introduction}

Human performance on many recognition tasks has been surpassed by deep neural networks~\cite{he_delving_2015, he_deep_2016} trained with large-scale supervised data~\cite{deng_imagenet:_2009, russakovsky_imagenet_2015} and specialized computational devices~\cite{jouppi_-datacenter_2017}. These trained neural networks, also known as pre-trained models, not only work well on tasks they are intended for but also produce generic representations~\cite{donahue_decaf:_2014} that benefit downstream tasks such as object detection~\cite{girshick_rich_2014}.

Apart from serving as fixed feature extractors, pre-trained models can be fine-tuned~\cite{, yosinski_how_2014, he_rethinking_2019} to serve downstream tasks better. The transfer learning paradigm ``pre-training $\rightarrow$ fine-tuning'' enjoys tremendous success in both vision~\cite{kornblith_better_2019} and language~\cite{devlin_bert:_2019} communities, and continues to expand to communities like geometric learning~\cite{hu_strategies_2020}. \emph{Transfer of pre-trained models has become one of the cornerstones of deep learning.}

\begin{figure}[htbp]
  \includegraphics[width=\columnwidth]{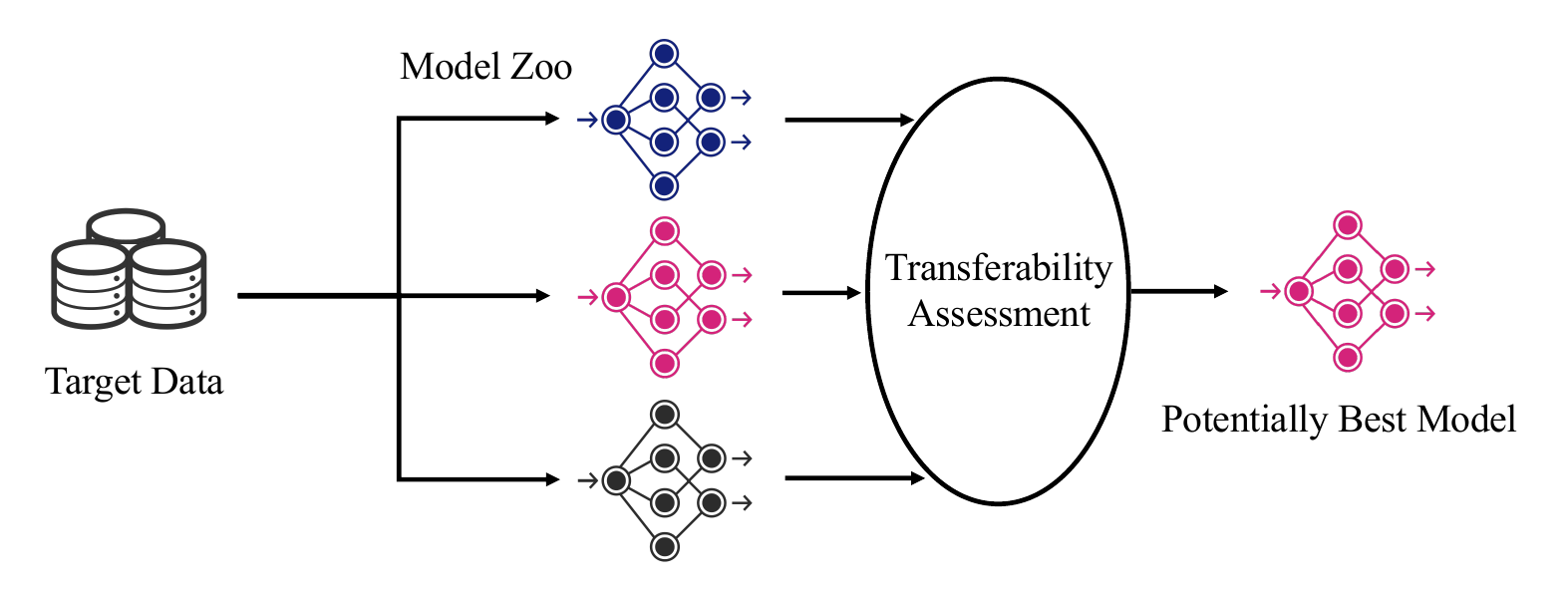}
  \vspace{-20pt}
  \caption{Illustration of task adaptive pre-trained model selection.}
  \label{fig:setting}
\end{figure}

Nowadays, there are numerous public pre-trained models offered by PyTorch~\cite{benoit_pytorch:_2019}, TensorFlow~\cite{abadi_tensorflow:_2016} and third-party libraries like HuggingFace Transformers~\cite{wolf_transformers:_2020}. When a practitioner wants to employ transfer learning to solve a specific task, the first problem is to select a good pre-trained model to start from. The problem is non-trivial and task adaptive, considering that \emph{different tasks favor different pre-trained models}. Figure~\ref{fig:setting} intuitively illustrates the problem.

The problem challenges researchers to develop a practical assessment method that is fast, accurate and general. It should be fast enough compared to brute-force fine-tuning all available pre-trained models~\cite{zamir_taskonomy:_2018}, should be accurate enough so that potentially best models can be identified, and should be general enough to tackle a wide variety of common learning scenarios. 

Despite its practical significance, there is limited guidance on task adaptive pre-trained model selection. Based on NCE~\cite{tran_transferability_2019}, \citet{nguyen_leep:_2020} recently studied the problem when both the pre-train task and the downstream task are classification. They construct an empirical predictor by estimating the joint distribution over the pre-trained and target label spaces and take the performance of the empirical predictor (LEEP) to assess pre-trained models. Though being fast, prior methods are not accurate and are specialized for transferring supervised pre-trained models to classification. They cannot apply to either contrastive pre-trained models~\cite{he_momentum_2020,chen_simple_2020}, unsupervised pre-trained language models~\cite{devlin_bert:_2019, liu_roberta:_2019}, or regression tasks.

Table~\ref{tab:applicability} shows the applicability of pre-trained model selection methods. Prior to this paper, for most ($4$ out of $5$) transfer learning settings, task adaptive pre-trained model selection does not have a decent solution.

\begin{table}[tbp]
  \centering
  \caption{\centering Applicability of prior methods and the proposed LogME. ``LM'' means language modeling.}
  \vspace{3pt}
  \resizebox{\columnwidth}{!}{
    \begin{tabular}{cccccc}
      \toprule
    Modality & Pre-train & Target & LEEP & NCE & LogME \\
    \midrule
    \multirow{4}[0]{*}{vision} & classification & classification &   \ding{51}    &   \ding{51}    & \ding{51} \\
          & classification & regression &  {\transparent{0.5} \ding{55}}     &  {\transparent{0.5} \ding{55}}     & \ding{51} \\
          & contrastive & classification &       {\transparent{0.5} \ding{55}}     &  {\transparent{0.5} \ding{55}}     & \ding{51}  \\
          & contrastive & regression &     {\transparent{0.5} \ding{55}}     &  {\transparent{0.5} \ding{55}}     & \ding{51} \\
          \hline
    language & LM & classification &       {\transparent{0.5} \ding{55}}     &  {\transparent{0.5} \ding{55}}     & \ding{51}  \\
    \bottomrule
    \end{tabular}}
  \label{tab:applicability}%
\end{table}%

To provide a general method for pre-trained model selection in various settings, we consider the features extracted by pre-trained models, thus being agnostic to how models are pre-trained. The maximum value of label evidence (marginalized likelihood) given extracted features is calculated, providing a general probabilistic approach that is applicable to both classification and regression tasks. Finally, the logarithm of maximum evidence (LogME) is used to assess pre-trained models for transfer learning. The maximum evidence is less prone to over-fitting~\cite{bishop_pattern_2006}, and its humongous computational cost is dramatically reduced by our carefully designed algorithm.

The contributions of this paper are two-fold:
\begin{itemize}
  \item We propose LogME for task adaptive pre-trained model selection, and develop a fast algorithm to accelerate the computation. LogME is easy to interpret and is extremely efficient. It brings at most $3000\times$ speedup in wall-clock time and requires just $1\%$ memory footprint, characterizing itself as the first practical method for assessing pre-trained models in various transfer learning settings.
  \item We extensively validate the generality and superior performance of LogME on \emph{$22$ pre-trained models} and \emph{$17$ downstream tasks}, covering various pre-trained models (supervised pre-trained and unsupervised pre-trained), downstream tasks (classification and regression), and modalities (vision and language).
\end{itemize}

\section{Related Works}

\subsection{Transfer learning}
\label{sec:transfer}

Transfer learning~\cite{thrun_learning_1998} is a broad research area containing transductive transfer, inductive transfer, task transfer learning, and so on. Transductive transfer is commonly known as domain adaptation~\cite{quionero-candela_dataset_2009, ganin_unsupervised_2015, long_learning_2015}, with the focus on eliminating domain shifts between two domains. Inductive transfer, or fine-tuning~\cite{erhan_why_2010, yosinski_how_2014}, leverages an inductive bias (a pre-trained model) to improve the performance on a target task and is extremely popular in deep learning. In task transfer learning~\cite{zamir_taskonomy:_2018}, researchers investigate how to transfer between tasks rather than pre-trained models. They aim to discover the relationship among tasks~\cite{ben-david_exploiting_2003} and to exploit the relationship for further development. In the context of deep learning, transfer learning usually refers to inductive transfer, the topic we are concerned about in this paper.

Besides the na\"ive fine-tuning where pre-trained models only serve as good initializations, there are sophisticated fine-tuning techniques like regularization~\cite{li_explicit_2018}, additional supervision~\cite{you_co-tuning_2020}, specially designed architecture~\cite{kou_stochastic_2020}, and intermediate-task training which continues to pre-train on an intermediate task~\cite{gururangan_dont_2020, pruksachatkun_intermediate-task_2020, garg_tanda:_2020}. They can improve transfer learning performance especially when the amount of target data is small, but in general, \emph{they do not change the ranking of pre-trained models in downstream tasks}. If pre-trained model A is better than pre-trained model B in a task with vanilla fine-tuning, typically A is still better than B when those sophisticated techniques are turned on. For example, on three datasets and four sampling rates from Table~2 in \citet{you_co-tuning_2020}, better fine-tuning performance mostly indicates better Co-Tuning (their proposed method) performance. Therefore we focus on vanilla fine-tuning rather than these techniques in the rest of the paper, but practitioners are encouraged to adopt them for further improvement after selecting a pre-trained model.

\subsection{Pre-trained models}

Pre-trained models are neural networks trained on large-scale datasets and can be transferred to downstream tasks. Popular pre-trained models are reviewed in the following.

\textbf{Supervised pre-trained models.} ImageNet is the most famous dataset for supervised pre-training. In the ImageNet classification challenge, \citet{he_delving_2015} developed the first deep neural network that surpassed human performance. InceptionNet~\cite{szegedy_going_2015} is another family of deep neural networks with parallel convolution filters. ResNet~\cite{he_deep_2016} introduces skip connections to ease the training and becomes much deeper with better performance. DenseNet~\cite{huang_densely_2017} has carefully designed densely-connected blocks. MobileNet~\cite{sandler_mobilenetv2:_2018} pays attention to mobile-friendly network structures, and the structure can be further optimized by network architecture search~\cite{tan_mnasnet:_2019}.

\textbf{Contrastive pre-trained models.} Although ImageNet pre-training is popular, the labeling cost of ImageNet is very high. Given the large amount of unlabeled data on the Internet, unsupervised pre-training has gained much attention in the past year. By exploiting self-supervised learning~\cite{jing_self-supervised_2020} on unlabeled data~\cite{mahajan_exploring_2018} with contrastive loss~\cite{gutmann_noise-contrastive_2010}, unsupervised contrastive pre-training produces a family of pre-trained models besides supervised pre-trained models. \citet{he_momentum_2020} proposed Momentum Contrast with a queue structure to fully exploit unlabeled data and obtained representations on par with supervised pre-training in terms of quality. \citet{chen_simple_2020} greatly improved the performance by exploring data augmentation, multi-layer projection head and many empirical design choices. How to design better contrastive pre-training strategies is still under active research~\cite{tian_what_2020}.

\textbf{Pre-trained language models.} In the language community, unsupervised pre-training has been well established by training masked language models~\cite{devlin_bert:_2019} or autoregressive language models~\cite{yang_xlnet:_2019} on a large unlabeled corpus. \citet{liu_roberta:_2019} explored many practical details on how to improve the training of these models. Because pre-trained language models are very large, \citet{sanh_distilbert_2019} proposed distillation to get smaller and faster models. These pre-trained language models become an indispensable component in winning submissions on common benchmarks like GLUE~\cite{wang_glue:_2018} and SQuAD~\cite{rajpurkar_squad:_2016}, and have profound industrial influence.

Pre-trained models are hosted in model zoos like \href{https://pytorch.org/vision/stable/models.html}{TorchVision} and \href{https://huggingface.co/models}{HuggingFace}. There are so many pre-trained models, but no one can overwhelmingly outperform the rest in all downstream tasks. The best model for a downstream task depends on the characteristic of both the task and the pre-trained model, thus being task adaptive. Practitioners can have a hard time choosing which pre-trained model to use for transfer learning, calling for a practical method to assess pre-trained models without brute-force fine-tuning.

\subsection{Assessing transferablitiy of pre-trained models}

Assessing transferability of pre-trained models has a great significance to guide common practice. \citet{yosinski_how_2014} studied which layer of a pre-trained model can be transferred while \citet{kornblith_better_2019} studied a wide variety of modern pre-trained models in computer vision. These papers aim for a deeper understanding of transfer learning~\cite{neyshabur_what_2020}. Nonetheless, they draw conclusions by expensive and exhaustive fine-tuning with humongous computation cost (Section~\ref{sec:timecost}) which is hard for practitioners to afford.

To \emph{efficiently} assess the transferability of pre-trained models, \citet{nguyen_leep:_2020} pioneered to develop LEEP with a focus on supervised pre-trained models transferred to classification tasks. The joint distribution over pre-trained labels and the target labels is estimated to construct an empirical predictor. The log expectation of the empirical predictor (LEEP) is used as a transferability measure. The LEEP method is closely related to Negative Conditional Entropy (NCE) proposed by \citet{tran_transferability_2019}, an information-theoretic quantity~\cite{cover_elements_1999} to study the transferability and hardness between classification tasks.

LEEP~\cite{nguyen_leep:_2020} and NCE~\cite{tran_transferability_2019}, the only two prior methods for pre-trained model selection, shed light on this problem but leave plenty of room for further performance improvement. In addition, they can only handle classification tasks with supervised pre-trained models. Since contrastive pre-training and language modeling tasks do not have categorical labels, \emph{prior methods cannot deal with these increasingly popular models}. To promote pre-trained model selection, we propose LogME which is broadly applicable to various pre-trained models, downstream tasks, and even data modalities.



\section{Problem setup}
\label{sec:setup}

In task adaptive pre-trained model selection, we are given $M$ pre-trained models $\{\phi_m\}_{m=1}^{M}$ and a target dataset $\mathcal{D} = \{(x_i, y_i)\}_{i=1}^{n}$ with $n$ labeled data points. The dataset has an evaluation metric (accuracy, MAP, MSE \emph{etc.}) to measure the ground-truth transfer performance $T_m$ of fine-tuning $\phi_m$ with proper hyper-parameter tuning. A practical assessment method should produce a score $S_m$ for each pre-trained model $\phi_m$ (ideally without fine-tuning $\phi_m$ on $\mathcal{D}$), and the scores $\{S_m\}_{m=1}^{M}$ should well correlate with $\{T_m\}_{m=1}^{M}$ so that top performing pre-trained models can be selected by simply evaluating the scores.

\paragraph{How to measure the performance of pre-trained model assessing methods.}

A perfect pre-trained model assessing method would output $\{S_m\}_{m=1}^{M}$ with exactly the same order as $\{T_m\}_{m=1}^{M}$. To measure the deviation from the perfect method, we can use simple metrics like top-1 accuracy or top-k accuracy (whether top-k in $\{S_m\}_{m=1}^{M}$ are also top-k in $\{T_m\}_{m=1}^{M}$). But top-1 accuracy is too conservative and top-k accuracy is not comparable across different values of $M$. Therefore we turn to rank correlation~\cite{fagin_comparing_2003} to directly measure the correlation between $\{S_m\}_{m=1}^{M}$ and $\{T_m\}_{m=1}^{M}$. The prior work~\cite{nguyen_leep:_2020} adopted Pearson's linear correlation coefficient, but neither Pearson's linear correlation nor its variant (Spearman's rank correlation) has a simple interpretation (see the interpretation of $\tau$ below).

Since the purpose of assessment is to choose a good pre-trained model, we hope \emph{$T_i$ is better than $T_j$ if $S_i$ is better than $S_j$}, which can be well captured by Kendall's $\tau$ coefficient~\citep{kendall_new_1938} as described in the following.

To simplify the discussion, assume larger value of transfer performance $T$ and score $S$ are preferred (\emph{e.g.} accuracy). If this is not the case (\emph{e.g.} transfer performance is measured by mean square error), the negation can be considered. For a pair of measures $(T_i, S_i)$ and $(T_j, S_j)$, the pair is concordant if $T_i < T_j \land S_i < S_j$ or $T_i > T_j \land S_i > S_j$ (concisely speaking, $\text{sgn}(T_i - T_j) \text{sgn}(S_i - S_j) = 1$). The Kendall's $\tau$ coefficient is defined by the following equation, which enumerates all $\binom{M}{2}$ pairs and counts the number of concordant pairs minus the number of discordant pairs.
\begin{equation*}
  \vspace{-3pt}
    \tau = \frac{2}{M(M-1)} \sum_{1 \le i<j \le M} \text{sgn}(T_i - T_j) \text{sgn}(S_i - S_j) 
\end{equation*}
\textbf{How to interpret $\tau$}~\cite{fagin_comparing_2003}. The range of $\tau$ is $[-1, 1]$. $\tau=1$ means $T$ and $S$ are perfectly correlated ($S_i > S_j \iff T_i > T_j$), and $\tau=-1$ means $T$ and $S$ are reversely correlated ($S_i > S_j \iff T_i < T_j$). \emph{If $T$ and $S$ have correlation of $\tau$, the probability of $T_i > T_j$ is $\frac{\tau + 1}{2}$ when $S_i > S_j$}.

\textbf{Pay attention to top performing models.} Since a major application of assessing pre-trained models is to select top performing pre-trained models, discordant / concordant pairs should be weighted more if $T_i, T_j, S_i, S_j$ are larger. This can be taken care of by $\tau_w$~\cite{vigna_weighted_2015}. The details of calculating $\tau_w$ can be found in \href{https://docs.scipy.org/doc/scipy/reference/generated/scipy.stats.weightedtau.html}{implementation} from SciPy~\cite{2020SciPy-NMeth}.

In short, we measure the correlation between $\{S_m\}_{m=1}^{M}$ and $\{T_m\}_{m=1}^{M}$ by the weighted variant $\tau_w$~\cite{vigna_weighted_2015}. \emph{Larger $\tau_w$ indicates better correlation and better assessment.}

Note that how to measure the performance of pre-trained model assessing methods is neither the focus nor the claimed novelty of this paper. We use weighted Kendall's $\tau$ because it is easy to interpret, but any proper rank correlation metric (such as Pearson's linear correlation and Spearman's rank correlation) can be adopted and should yield similar conclusions on superiority of our proposed method.

\section{The LogME approach}
\label{sec:logme}

For each pre-trained model $\phi_m$, the algorithm should produce a score $S_m$ independent from the rest of pre-trained models. We thus drop the subscript $m$ in this section.

To be fast, we try to \emph{avoid gradient optimization}. The pre-trained model $\phi$ serves as a fixed feature extractor. Features $\{f_i = \phi(x_i)\}_{i=1}^{n}$ and labels $\{y_i\}_{i=1}^{n}$ are used to assess pre-trained models. Note that \citet{nguyen_leep:_2020} used a pre-trained classification head $h$ besides the pre-trained representation model $\phi$, limiting their method to supervised pre-trained models. In contrast, \emph{we only use the pre-trained representation model $\phi$ so that the proposed method can be applied to any pre-trained model} (whether supervised pre-trained or unsupervised pre-trained).

Without gradient optimization, the problem is cast into estimating the compatibility of features $\{f_i = \phi(x_i)\}_{i=1}^{n}$ and labels $\{y_i\}_{i=1}^{n}$, which is discussed in the rest of this section.

\subsection{Evidence calculation}
\label{sec:evidence}

We first consider a simple case, with features $f_i \in \mathbb{R}^D$ and scalar labels $y_i \in \mathbb{R}$. The feature matrix $F \in \mathbb{R}^{n \times D}$ contains all the features and $y \in \mathbb{R}^{n}$ denotes all the labels.

A direct measurement of the compatibility between features $F$ and labels $y$ is the probability density $p(y | F)$, which is intractable without a parametrized model. Since \emph{the rule-of-thumb transfer learning practice is to add a fully-connected layer on top of the pre-trained model}, we use a linear model upon features parametrized by $w$.

\begin{figure}[htbp]
  \centering
  \includegraphics[width=1\columnwidth]{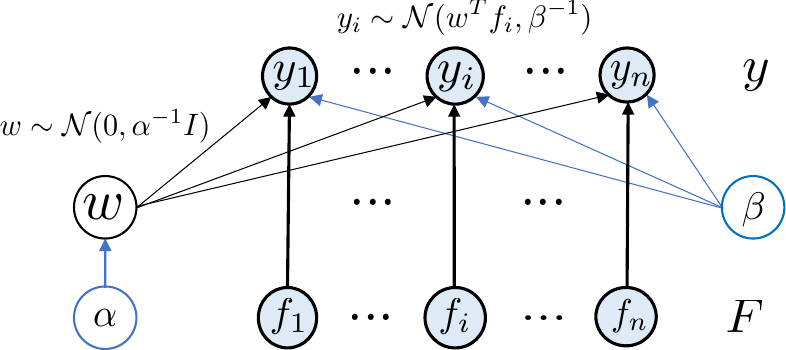}
  \vspace{-15pt}
  \caption{The directed graphical model for calculating evidence.}
  \label{fig:graph}
\end{figure}

A na\"ive approach to deal with the linear model is to find the best $w^*$ by logistic / linear regression and to assess pre-trained models by likelihood $p(y | F, w^*)$. However, it is well-known that \textit{likelihood is prone to over-fitting}~\cite{bishop_pattern_2006}, which is experimentally observed in Supplementary~B. A better approach is to use the evidence (marginalized likelihood) $p(y|F) = \int p(w) p(y | F, w) \mathrm{d} w$, which integrates over all possible values of $w$ and is better than simply using one optimal value $w^*$. This evidence-based approach is an elegant model selection approach and has a rigorous theoretical foundation~\cite{knuth_bayesian_2015}. For $p(w)$ and $p(y | F, w)$, we use the commonly adopted graphical model (Figure~\ref{fig:graph}) specified by two positive parameters $\alpha$ and $\beta$: the prior distribution of the weight is an isotropic multivariate Gaussian $w \sim \mathcal{N}(0, \alpha^{-1}I)$, and the distribution of each observation is a one-dimensional normal distribution $p(y_i | f_i, w, \beta) = \mathcal{N}(y_i | w^T f_i, \beta^{-1})$.

According to the causal structure in Figure~\ref{fig:graph} and the basic principles in graphical models~\cite{koller_probabilistic_2009}, the evidence can be calculated analytically as Eq.~\ref{eq:likelihood}.
\begin{equation}
  \begin{aligned}
    p(y|F, \alpha, \beta) &= \int p(w | \alpha) p(y | F, w, \beta) \mathrm{d} w \\
    &= \int p(w | \alpha) \prod_{i=1}^{n} p(y_i | f_i, w, \beta) \mathrm{d} w \\
    &= (\frac{\beta}{2 \pi})^{\frac{n}{2}} (\frac{\alpha}{2 \pi})^{\frac{D}{2}} \int e^{-\frac{\alpha}{2} w^Tw -\frac{\beta}{2} \vert \vert Fw - y \vert \vert^2}  \mathrm{d} w
  \end{aligned}
  \label{eq:likelihood}
\end{equation}
As $\int e^{- \frac{1}{2} (w^TAw + b^Tw + c)} \, \mathrm{d}w = \sqrt{\frac{(2\pi)^D}{\vert A \vert}} e^{-\frac{1}{2}c + \frac{1}{8}b^TA^{-1}b}$ when $A$ is positive definite, Eq.~\ref{eq:likelihood} can be simplified. By taking the logarithm to make the equation simple, Eq.~\ref{eq:evidence} shows the logarithm of the evidence as a function of $\alpha, \beta$, where $A = \alpha I + \beta F^TF, m = \beta A^{-1}F^Ty$.
\begin{equation}
  \begin{aligned}
    \mathcal{L}(\alpha, \beta) &=  \log p(y|F, \alpha, \beta) \\ &= \frac{n}{2} \log \beta  + \frac{D}{2} \log \alpha - \frac{n}{2} \log 2\pi \\
    & - \frac{\beta}{2} \vert \vert F m - y \vert \vert_2^2 - \frac{\alpha}{2} m^Tm - \frac{1}{2} \log \vert A \vert
  \end{aligned}
  \label{eq:evidence}
\end{equation}

\subsection{Evidence maximization and LogME}

A remaining issue of Eq.~\ref{eq:evidence} is how to determine $\alpha, \beta$. \citet{gull_developments_1989} suggested that we should choose $\alpha, \beta$ to maximize the evidence, \emph{i.e.} use $(\alpha^*, \beta^*) = \arg \max_{\alpha, \beta} \mathcal{L}(\alpha, \beta)$. Because $m$ and $A$ are coupled, maximizing $\mathcal{L}(\alpha, \beta)$ is generally a difficult problem. However, this form of maximization can be achieved by alternating between evaluating $m, \gamma$ and maximizing $\alpha, \beta$ with $m, \gamma$ fixed~\citep{gull_developments_1989}, resulting the following formula, where $\sigma_i$'s are singular values of $F^TF$.
\begin{equation*}
  A = \alpha I + \beta F^TF, m = \beta A^{-1}F^Ty, \gamma = \sum_{i=1}^{D} \frac{\beta \sigma_i}{\alpha + \beta \sigma_i}
  \label{eq:fix1}
\end{equation*}
\begin{equation*}
    \alpha \leftarrow \frac{\gamma}{m^Tm}, \beta \leftarrow \frac{n - \gamma}{\vert \vert F m - y \vert \vert_2^2}
    \label{eq:fix2}
\end{equation*}
When the fixed-point iteration converges (empirically it converges with no more than three iterations), the logarithm maximum evidence $\mathcal{L}(\alpha^*, \beta^*)$ is used to evaluate the compatibility between features and labels. Because $\mathcal{L}(\alpha^*, \beta^*)$ scales linearly with $n$, we normalize it by $\frac{\mathcal{L}(\alpha^*, \beta^*)}{n}$ and term it LogME (logarithm of of maximum evidence). It can be intuitively interpreted as the average maximum log evidence of labels given the pre-trained features.

\textbf{Extending LogME to complex cases.} The LogME approach described above starts from a single-target regression. If the target problem is a multivariate-regression task, \emph{i.e.} $y \in \mathbb{R}^{n\times K}$, we can calculate LogME for each dimension $k \; (1 \le k \le K)$ and average them over the $K$ dimension. If the target problem is a classification task with $K$ classes, Eq.~\ref{eq:likelihood} cannot be calculated analytically~\cite{daunizeau_semi-analytical_2017} with a categorical prior distribution, but we can convert the labels to one-hot labels and treat the problem as multivariate regression. Therefore, LogME can be used in both classification and regression tasks. The overall algorithm of LogME is described in Algorithm~\ref{alg:logme}.

\begin{algorithm}[htbp]
	\caption{LogME}
	\label{alg:logme}
	\begin{algorithmic}[1]
		\STATE {\bfseries Input:} Pre-trained model $\phi$ \\	
		\quad  \quad \;\;\; {Target dataset} $\mathcal{D} = \{(x_i, y_i)\}_{i=1}^{n}$  \\ 
		\STATE {\bfseries Output:} logarithm of maximum evidence (LogME)
		
		\vspace{5pt}
		
		\STATE { Extract} features using pre-trained model $\phi$:
		
    \quad \quad \quad \quad $F \in \mathbb{R}^{n\times D}$, $f_i = \phi(x_i)$, $Y \in \mathbb{R}^{n\times K}$
    		
    \STATE{Compute} SVD $F^TF = V\text{diag}\{\sigma\}V^T$ \\
    
    \FOR{$k=1$ \text{to} $K$ } 
    \STATE{Let} $y = Y^{(k)} \in \mathbb{R}^n$, initialize $\alpha=1, \beta=1$ \\ 
      \WHILE{$\alpha, \beta$ not converge}
      \vspace{3pt}
        \STATE{Compute} $\gamma = \sum_{i=1}^{D} \frac{\beta \sigma_i}{\alpha + \beta \sigma_i}, \Lambda = \text{diag}\{(\alpha + \beta \sigma)\}$ \\
        \vspace{3pt}
        \STATE{\textbf{Na\"ive}}: $A = \alpha I + \beta F^TF, m = \beta A^{-1}F^Ty$ \\
        \vspace{3pt}
        \STATE{\textbf{Optimized}}: $m = \beta (V (\Lambda^{-1} (V^T (F^T y))))$ \\
        \vspace{3pt}
        \STATE{Update} $\alpha \leftarrow \frac{\gamma}{m^Tm}, \beta \leftarrow \frac{n - \gamma}{\vert \vert F m - y \vert \vert_2^2}$  \\
      \ENDWHILE
      \STATE {Compute} $\mathcal{L}_k = \frac{1}{n} \mathcal{L}(\alpha, \beta)$ using Eq.~\ref{eq:evidence} 
    \ENDFOR
		
		\STATE{Return} LogME $\frac{1}{K}\sum_{k=1}^{K} \mathcal{L}_k$ \\
		
	\end{algorithmic}
\end{algorithm}

\subsection{Computational speedup}
\label{sec:speedup}

Although the Bayesian approach of maximum evidence has many nice properties~\cite{knuth_bayesian_2015}, it inherits the common drawback of Bayesian methods with high computational complexity. The na\"ive implementation of Algorithm~\ref{alg:logme} has a complexity of $\mathcal{O}(KD^3 + nKD^2)$. For typical usage with $D \approx 10^3, n \approx 10^4, K \approx 10^3$, the computational cost is $10^{13}$, making the wall-clock time comparable to fine-tuning the pre-trained model $\phi$.

Notice that the most expensive operations are Line~$9$ with matrix inversion $A^{-1}$ and matrix multiplication $A^{-1}F^T$. These expensive operations, however, can be avoided by exploiting the decomposition of $F^TF$, which is readily accessible from Line~4.

To avoid matrix inversion $A^{-1}$, we exploit the decomposition $F^TF = V \text{diag}\{\sigma\} V^T$ ($V$ is an orthogonal matrix). Let $\Lambda = \text{diag}\{(\alpha + \beta \sigma)\}$, then $A = \alpha I + \beta F^TF = V\Lambda V^T$, and $A^{-1} = V\Lambda^{-1}V^T$. To avoid the matrix-matrix multiplication $A^{-1}F^T$, we notice that $y$ is a column vector and the associative law admits a fast computation $A^{-1} F^T y = (V (\Lambda^{-1} (V^T (F^T y))))$. In each for-loop, we only need to update $\Lambda$ rather than the expensive $A^{-1}$. In this way, all matrix-matrix multiplications are reduced to matrix-vector product, and the matrix inversion is avoided, as described in Line~$10$. Table~\ref{tab:optimization} analyzes the complexity in detail. The optimized algorithm makes a time-consuming Bayesian approach fast enough, reducing the wall-clock time by the order of $10^2$ (see Section~\ref{sec:timecost}).

\begin{table}[htbp]
  \centering
  \caption{Computational complexity of Algorithm~\ref{alg:logme}.}
  \vspace{3pt}
  \resizebox{\columnwidth}{!}{
    \begin{tabular}{ccc}
      \toprule
          & Complexity per for-loop & Overall complexity \\
          \midrule
    na\"ive &   $\mathcal{O}(D^3 + n D^2)$    & $\mathcal{O}(KD^3 + nKD^2)$ \\
    optimized &   $\mathcal{O}(D^2 + n D)$    & $\mathcal{O}(KD^2 + nKD + D^3 + n D^2)$ \\
    \bottomrule
    \end{tabular}}
  \label{tab:optimization}%
\end{table}%

The proposed LogME is easy to interpret, has a solid theoretical foundation, and is applicable to various settings. Its computational cost is dramatically reduced by our optimized implementation.

\section{Experiments}

We first present the illustration of LogME on toy problems, and then focus on task adaptive pre-trained model selection. Original data are available in Supplementary~C.

\paragraph{Illustration with toy data.}

To give readers an intuitive sense of how LogME works, we generate features with increasing noise to mimic the features extracted by pre-trained models with decreasing transferability and to check if LogME can measure the quality of features.

\begin{figure}[htbp]
  \centering
  \includegraphics[width=.98\columnwidth]{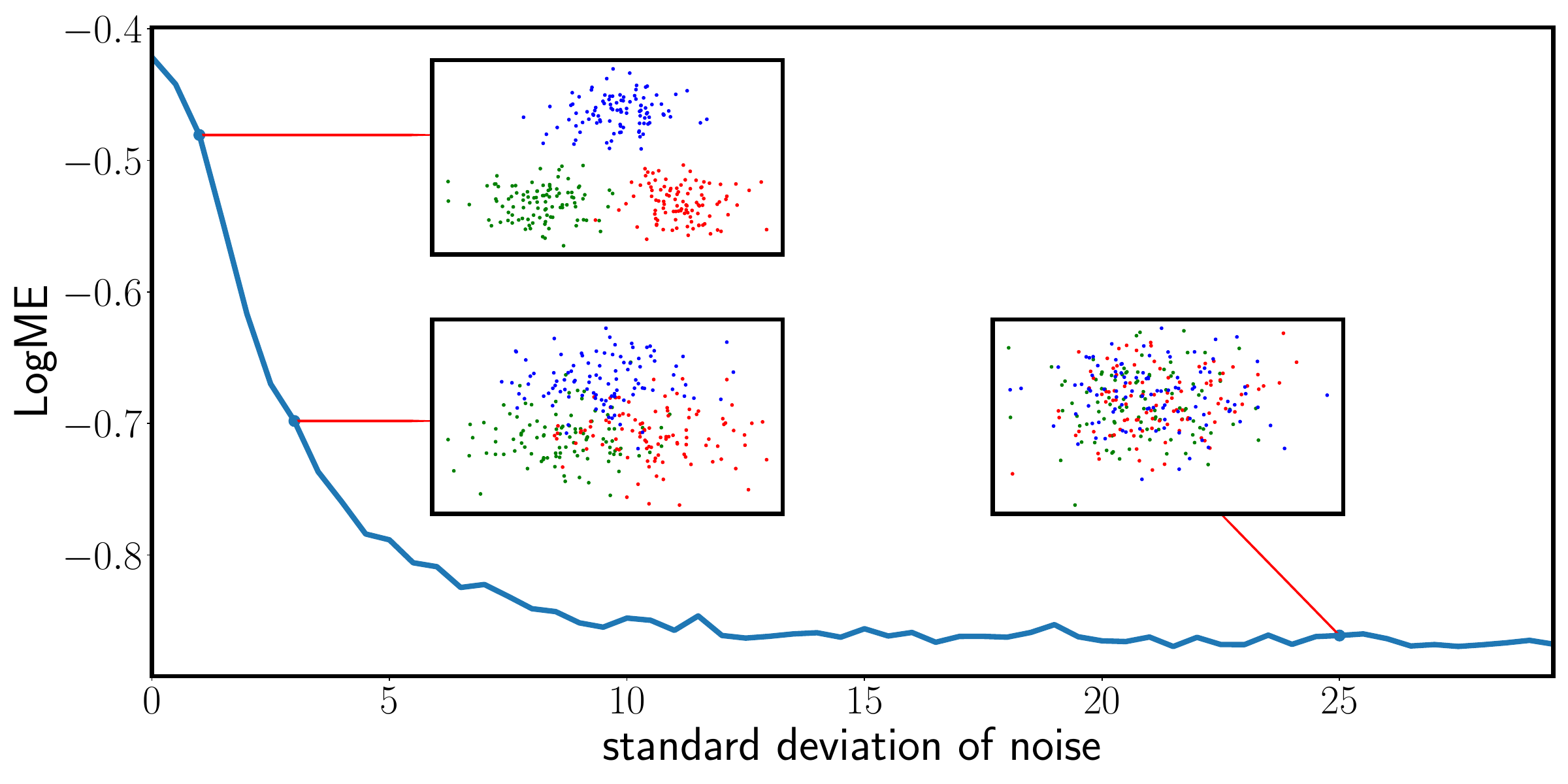}
  \includegraphics[width=.98\columnwidth]{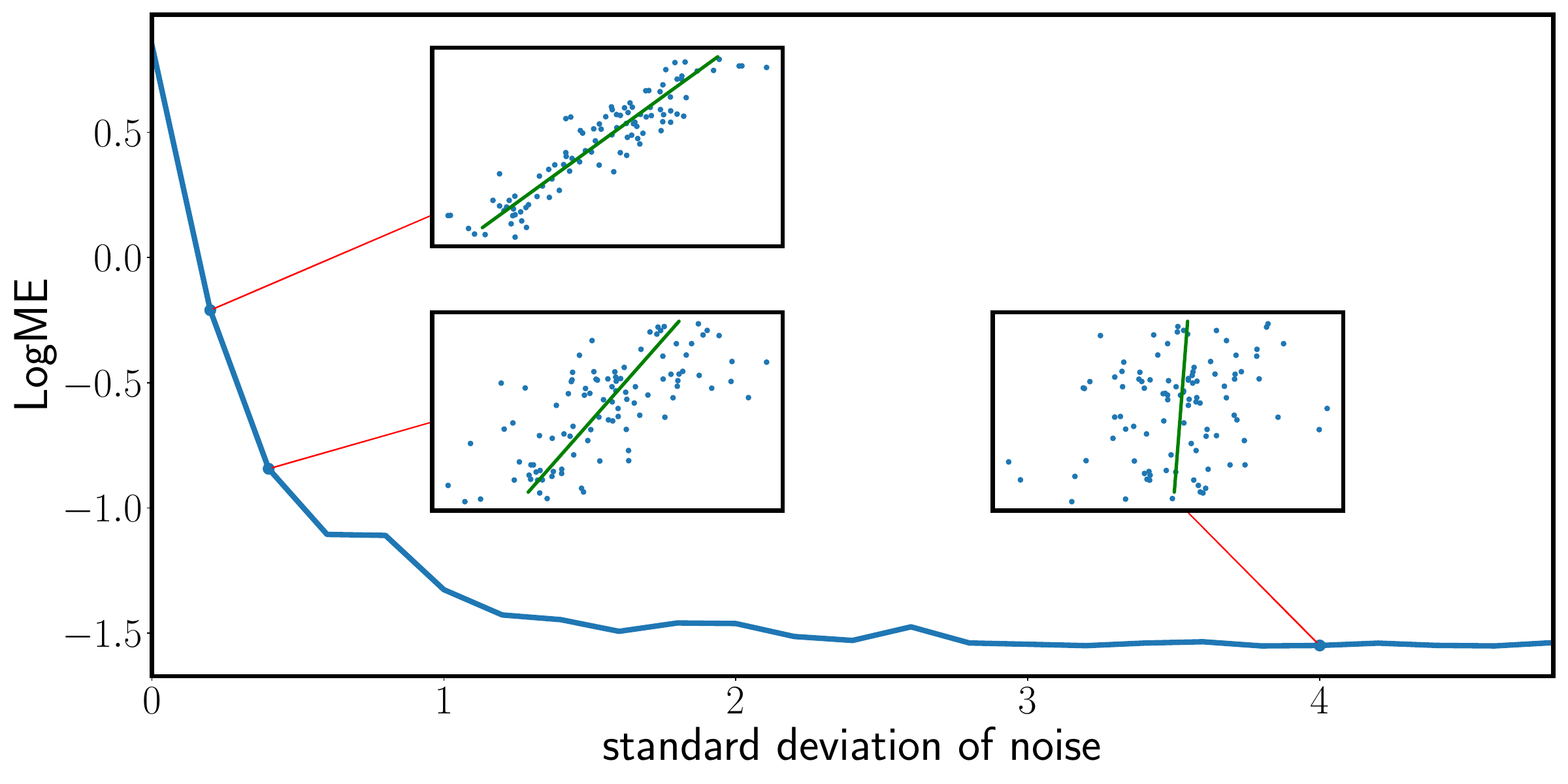}
  \caption{Illustration of LogME with toy data. It is clear that LogME decreases with decreasing feature quality.}
  \label{fig:toy}
\end{figure}

For classification (Figure~\ref{fig:toy} top), three clusters in 2-D plane are generated, with colors indicating the categories. Initially, the features are separable so LogME has a large value. Then we add Gaussian noise with increasing variance and LogME becomes smaller as expected.

For regression (Figure~\ref{fig:toy} bottom), $x$ is uniformly distributed and the output $y = 2 x + \epsilon$ with observation error $\epsilon \sim \mathcal{N}(0, 0.1^2)$. By adding noise to the feature $x' = x + \mathcal{N}(0, t^2)$, the quality of feature $x'$ becomes worse and it is harder to predict $y$ from $x'$. With larger $t$ (the standard deviation of noise), LogME becomes smaller as expected.

These toy experiments on synthesized data shows that LogME is a good measure of the feature quality, and therefore can provide a general assessment of pre-trained models for transfer learning.


\subsection{Transferring supervised pre-trained models to classification tasks}
\label{sec:supervised_classification}

We use $10$ ImageNet pre-trained models available from PyTorch: Inception~V1~\cite{szegedy_going_2015}, Inception~V3~\cite{szegedy_rethinking_2016}, ResNet~50~\cite{he_deep_2016}, ResNet~101~\cite{he_deep_2016}, ResNet~152~\cite{he_deep_2016}, DenseNet~121~\cite{huang_densely_2017}, DenseNet~169~\cite{huang_densely_2017}, DenseNet~201~\cite{huang_densely_2017}, MobileNet~V2~\cite{sandler_mobilenetv2:_2018}, and NASNet-A Mobile~\cite{tan_mnasnet:_2019}. These pre-trained models cover most of the supervised pre-trained models in transfer learning that practitioners frequently use.

For downstream classification tasks, we take $9$ commonly used datasets: Aircraft~\cite{maji_fine-grained_2013}, Birdsnap~\cite{berg_birdsnap:_2014}, Caltech~\cite{fei-fei_learning_2004}, Cars~\cite{krause_collecting_2013}, CIFAR10~\cite{krizhevsky_learning_2009}, CIFAR100~\cite{krizhevsky_learning_2009}, DTD~\cite{cimpoi_describing_2014}, Pets~\cite{parkhi_cats_2012}, and SUN~\cite{xiao_sun_2010}. Due to space limit, we leave the description of each dataset and data statistics in Supplementary~A.

\begin{figure*}[htbp]
  \begin{center}
    \includegraphics[width=\textwidth]{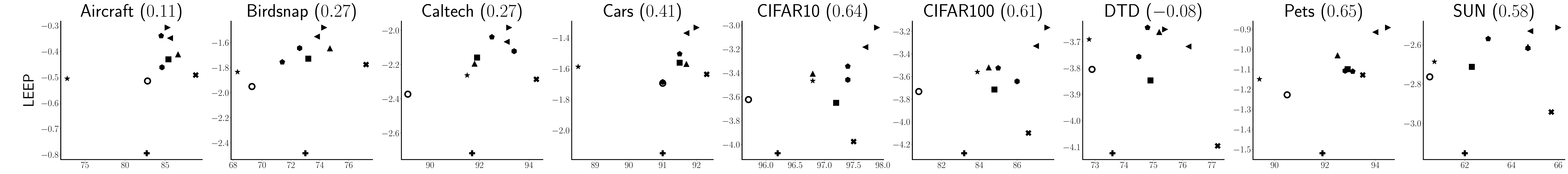}
  \includegraphics[width=\textwidth]{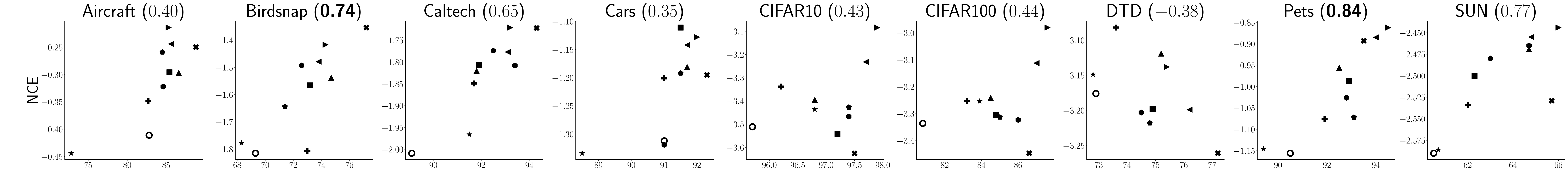}
  \includegraphics[width=\textwidth]{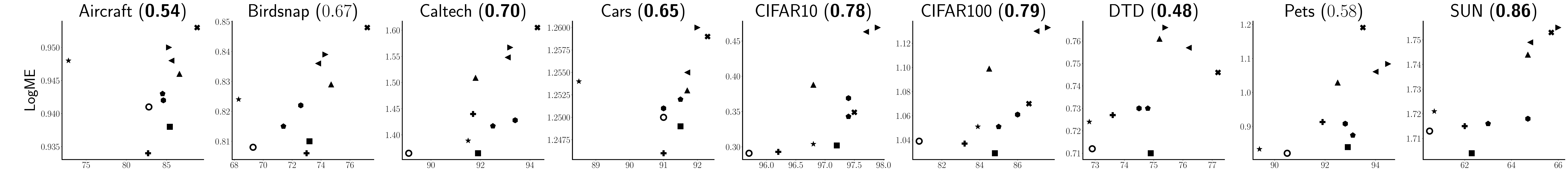}
  \end{center}
  \hfill \includegraphics[width=.96\textwidth]{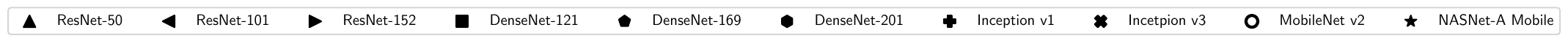}
  \caption{Correlation ($\tau_w$) between fine-tuned accuracy (X-axis) and three methods for pre-trained model selection on 9 datasets with 10 pre-trained models. One row for each method, one column for each dataset (with $\tau_w$ in the bracket near the dataset name), and one marker for each pre-trained model. The best $\tau_w$ in each dataset is marked in bold.}
  \label{fig:pre_train_selection}
\end{figure*}

To compute the value of transfer performance $\{T_m\}_{m=1}^{M}$ ($M = 10$), we carefully fine-tune pre-trained models with grid-search of hyper-parameters. As pointed out by \citet{li_rethinking_2020}, learning rates and weight decays are the two most important hyper-parameters. Hence we grid search learning rates and weight decays (7 learning rates from $10^{-1}$ to $10^{-4}$, 7 weight decays from $10^{-6}$ to $10^{-3}$, all logarithmically spaced) to select the best hyper-parameter on the validation set and compute the accuracy on the test set. \emph{It is noteworthy that LogME requires neither fine-tuning nor grid search.} Here we fine-tune pre-trained models to evaluate LogME itself, but practitioners can straightforwardly use LogME to evaluate pre-trained models without fine-tuning.

We compare LogME against LEEP~\cite{nguyen_leep:_2020} and NCE~\cite{tran_transferability_2019}. Prior to this paper, LEEP and NCE are the only two methods for pre-trained model selection without fine-tuning, and they are dedicated to transferring supervised pre-trained models to classification tasks. We use LEEP, NCE and LogME to compute scores $\{S_m\}_{m=1}^M$ by applying $10$ pre-trained models to the datasets. The correlation $\tau_w$ between scores and fine-tuned accuracies are presented in Figure~\ref{fig:pre_train_selection}. 

We can find that LogME has consistently better correlation than LEEP, and outperforms NCE on most datasets (7 datasets out of 9 datasets). Note that LEEP and NCE even show a negative correlation in DTD~\cite{cimpoi_describing_2014}, because they rely on the relationship between classes of the pre-trained task and the target task while DTD classes are very different from ImageNet categories. In contrast, LogME still performs reasonably well for DTD.

The smallest $\tau_w$ of LogME in Figure~\ref{fig:pre_train_selection} is around $0.5$, so the probability of a pre-trained model $\phi_1$ transferring better than $\phi_2$ is at least $75\%$ if  $\phi_1$ has a larger LogME. For most tasks $\tau_w$ of LogME is $0.7$ or $0.8$, so the probability of correct selection is $85\%$ or $90\%$, sufficient for practical usage.

\subsection{Transferring supervised pre-trained models to a regression task}

Besides extensive classification tasks considered above, this section shows how LogME can be used to assess pre-trained models for a regression task, while prior methods (LEEP and NCE) cannot.

The regression task we use is dSprites~\cite{matthey_dsprites:_2017} from VTAB~\cite{zhai_large-scale_2020} which is commonly used for evaluating the quality of learned representations. The input is an image containing a sprite (heart, square, and ellipse) with varying scale, orientation, and position. Pre-trained models are transferred to predict four scalars (scale, orientation, and $(x, y)$ positions) together, and mean square error (MSE) on the test data is reported. The supervised pre-trained models are the same as Section~\ref{sec:supervised_classification} and hyper-parameter tuning scheme follows.

\begin{figure}[htbp]
  \centering
  \includegraphics[width=\columnwidth]{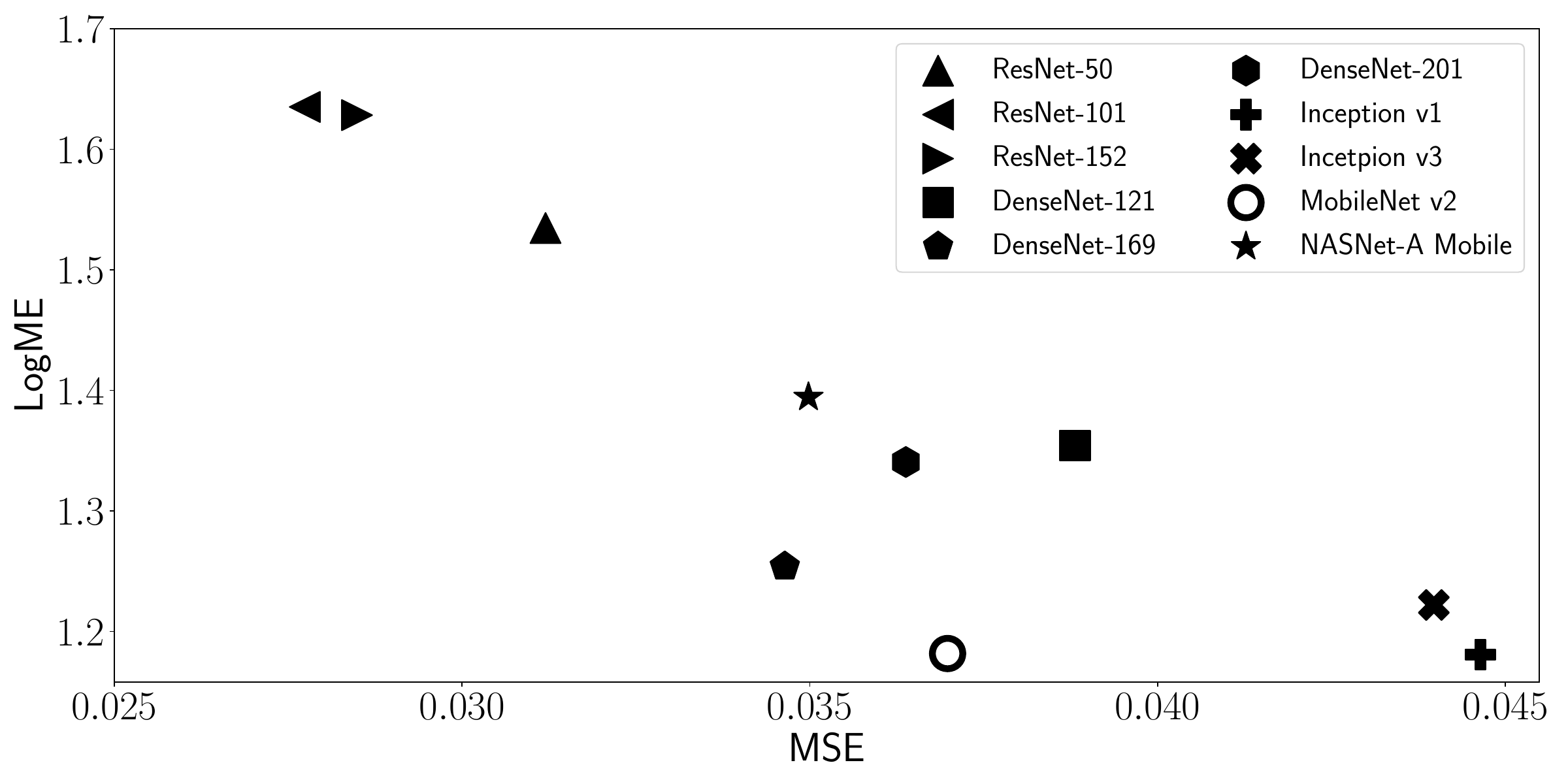}
  \vspace{-20pt}
  \caption{Supervised pre-trained models transferred to dSprites.}
  \label{fig:regression}
\end{figure}

Results are plotted in Figure~\ref{fig:regression}. It is clear that LogME and MSE are well correlated and the correlation coefficient $\tau_w = 0.84$ is very large: if a pre-trained model $\phi_1$ has larger LogME than $\phi_2$, with $92\%$ probability $\phi_1$ is better (has smaller MSE) than $\phi_2$ after actually fine-tuning.

\begin{figure*}[htbp]
  \centering
  \includegraphics[width=\textwidth]{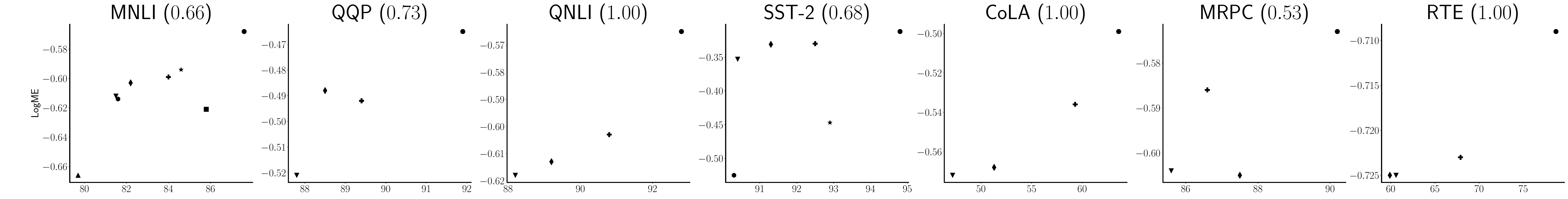}
  \includegraphics[width=.5\textwidth]{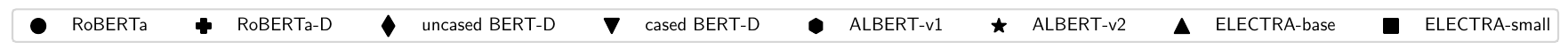}
  \caption{Correlation ($\tau_w$) between fine-tuned accuracy (X-axis) and LogME in 7 GLUE tasks with 8 popular pre-trained language models. One column for each task (with $\tau_w$ in the bracket near the task name), and one marker for each pre-trained model.}
  \label{fig:nlp}
\end{figure*}

\subsection{Transferring contrastive pre-trained models to downstream tasks}

The recently emerging unsupervised pre-trained models~\cite{he_momentum_2020} have a projection head with continuous output. However, LEEP and NCE cannot be extended to deal with the projection head of contrastive-based unsupervised pre-trained models because they rely on the relationship between pre-training categories and target categories.

Since LogME only requires features extracted from pre-trained models, it can be applied to contrastive pre-trained models. To demonstrate this, we use three popular models pre-trained with various training scheme: MoCo~V1~\cite{he_momentum_2020} with momentum contrast, MoCo~V2~\cite{chen_improved_2020} with an MLP projection head and strong data augmentation, MoCo~800 trained with 800 epochs as suggested by \citet{chen_simple_2020}, and SimCLR~\citep{chen_simple_2020} with carefully designed implementation.

Aircraft~\cite{maji_fine-grained_2013}, the first dataset (alphabetically) in Section~\ref{sec:supervised_classification} is used as the classification task, and dSprites~\cite{matthey_dsprites:_2017} is used as the regression task. Results are shown in Table~\ref{tab:unsupervised}. SimCLR on dSprites is not reported because it does not converge after several trials. LogME gives the \emph{perfect order} of both transferred accuracy and MSE. Note that the order in Aircraft (MoCo~V1 $<$ MoCo~V2 $<$ MoCo 800) is different from the order in dSprites (MoCo~V1 $<$ MoCo 800 $<$ MoCo~V2), so the transfer learning performance depends on both the pre-trained model and the target data, emphasizing the importance of \emph{task adaptive} pre-trained model selection. We also observe that LogME values of unsupervised pre-trained models are similar, mainly because unsupervised features are not very discriminative.

\begin{table}[htbp]
  \centering
  \caption{Use LogME to assess unsupervised pre-trained models.}
  \vspace{3pt}
  \resizebox{\columnwidth}{!}{
    \begin{tabular}{ccccc}
      \toprule
    \multirow{2}[0]{*}{Pre-trained Network} & \multicolumn{2}{c}{Aircraft} & \multicolumn{2}{c}{dSprites} \\
    \cmidrule(lr){2-3} \cmidrule(lr){4-5}
          &   Accuracy (\%)   &   LogME    &  MSE     & LogME \\
          \midrule
          MoCo V1 &   81.68    &   0.934    &   0.069    &  1.52 \\
          MoCo V2 &    84.16   &    0.941   &    0.047   &  1.64 \\
          MoCo 800 &    86.99   &    0.946   &   0.050    &  1.58 \\
          SimCLR &    88.10   &    0.950   &   -    &  - \\
          \midrule
          &     \multicolumn{2}{c}{$\tau_w$: 1.0}     &     \multicolumn{2}{c}{$\tau_w$: 1.0}  \\
          \bottomrule
    \end{tabular}}
  \label{tab:unsupervised}
\end{table}%

\subsection{Transferring pre-trained language models to the GLUE benchmark}

To further demonstrate the generality of LogME, we show how LogME can work for pre-trained language models. Again prior works (LEEP and NCE) cannot deal with these pre-trained language models.

Here we take an alternative approach of evaluating the transfer performance $\{T_m\}_{m=1}^M$. We do not fine-tune pre-trained models ourselves, but directly use accuracies tuned by others, and check if LogME can correlate well with the results. The HuggingFace \href{https://huggingface.co/models}{Model Hub} generously provides lots of pre-trained language models and even provides carefully tuned transfer learning results in some GLUE~\cite{wang_glue:_2018} tasks for some models. We take out pre-trained models that have GLUE performance tuned by the HuggingFace organization, and select the top $8$ downloaded models: RoBERTa~\cite{liu_roberta:_2019}, RoBERTa-D, uncased BERT-D, cased BERT-D, ALBERT-v1~\cite{lan_albert:_2020}, ALBERT-v2~\cite{lan_albert:_2020}, ELECTRA-base~\cite{clark_electra:_2020}, and ELECTRA-small~\cite{clark_electra:_2020} (``D'' means distilled version). The LogME on seven GLUE classification tasks together with fine-tuned accuracy are plotted in Figure~\ref{fig:nlp}. Some models only have results for certain tasks and we keep them as they are. Even though these accuracy numbers are tuned by the HuggingFace organization, \emph{LogME perfectly estimates the ranking of transfer performance for $3$ tasks} (with $\tau_w = 1$), showing the surprising effectiveness of LogME in pre-trained model selection.

\begin{table*}[!htbp]
  \centering
  \caption{Efficiency of LogME.}
  \vspace{3pt}
    \begin{tabular}{c|l|l}
          \toprule
          & wall-clock time & memory footprint \\
          \hline
    \multirow{4}[0]{*}{Computer Vision} &  fine-tune (upper bound) \hfill $161000$s    & fine-tune (upper bound) \hfill 6.3 GB\\
    &    extract feature (lower bound) \hfill $37$s  & extract feature (lower bound) \hfill 43 MB \\
          &    LogME \hfill $50$s  & LogME \hfill 53 MB \\
          &     benefit \hfill $3200\uparrow$ & benefit  \hfill $120\uparrow$  \\
          \hline
    \multirow{4}[0]{*}{Natural Language Processing} &      fine-tune (upper bound)  \hfill 100200s   & fine-tune (upper bound)\hfill 88 GB \\
    &    extract feature (lower bound) \hfill $1130$s  & extract feature (lower bound) \hfill 1.2 GB \\

          &    LogME \hfill 1157s  & LogME \hfill 1.2 GB \\
          &    benefit \hfill $86\uparrow$  &  benefit \hfill $ 73 \uparrow$\\
          \bottomrule
    \end{tabular}%
  \label{tab:efficiency}
\end{table*}%

\subsection{Efficiency of LogME}
\label{sec:timecost}

LogME is a practical method to assess pre-trained models for transfer learning because it is general, accurate, and efficient. Section~\ref{sec:logme} shows the generality of LogME by considering features and labels in the general form. Results in this section validates the strong correlation between LogME and ground-truth transfer learning performance, demonstrating that LogME is accurate. Next we quantitatively measure the efficiency of LogME compared to brute-force fine-tuning. The algorithmic complexity is presented in Section~\ref{sec:speedup}, thus we focus on wall-clock time and memory footprint here.

Results are shown in Table~\ref{tab:efficiency}. ResNet~50 on Aircraft is used for computer vision, and RoBERTa-D on MNLI task is used for NLP. Both wall-clock time and memory footprint is reported. The cost of computing ground-truth transferability $T_m$ (fine-tuning with hyper-parameter search) serves as the upper bound of pre-trained model assessment. We also list the cost of extracting features by pre-trained models as a reference, which is the lower bound of pre-trained model assessment. The cost for the rest models and datasets vary, but the proportion is similar. Note that, because carelessly tuned hyper-parameters cannot tell good models from bad models, it is necessary to attribute the cost of hyper-parameter search to brute-force fine-tuning while LogME does not need hyper-parameter tuning.

It is clear that brute-force fine-tuning is computationally expensive, requiring about a day for one dataset with one pre-trained model. Selecting the best pre-trained model out of $10$ models would cost $10$ days. Extracting features is very cheap and costs much less. In computer vision, the wall-clock time of LogME is reduced dramatically to $0.31 \permil $ of fine-tuning, bringing over $3000\times$ speedup while requiring $120\times$ less memory footprint. In the NLP domain, feature extraction is much slower and therefore the wall-clock time speedup is not as striking as computer vision, but still reaching $86\times$ speedup. \emph{In all cases, LogME costs almost the same as the lower bound (feature extraction)}, meaning that LogME makes practical assessment possible with minimal additional cost.

\section{Conclusion}

A fast, accurate, and general assessment of pre-trained models for transfer learning has great practical significance. This paper takes a probabilistic approach and proposes logarithm of maximum evidence (LogME) to tackle the task adaptive pre-trained model selection problem. The expensive computation of maximizing the marginalized likelihood is optimized by careful implementation, leading to over $3000\times$ speedup compared to vanilla fine-tuning. LogME is applicable to vast transfer learning settings with supervised pre-trained models and unsupervised pre-trained models, downstream classification and regression tasks, vision and language modalities. The impressive generality of LogME and its substantially better performance over prior methods can be interesting to many practitioners.

This paper measures the quality of pre-trained models by their static representations (\emph{i.e.} representations before fine-tuning). It is interesting to consider the dynamic representations (\emph{i.e.} representations after fine-tuning) of pre-trained models to account for the change of pre-trained models during fine-tuning. We leave it as a future work.

\section*{Acknowledgements}

We would like to thank Ximei Wang, Xinyang Chen, Yang Shu, and Yonglong Tian for helpful discussions.
This work was supported by the National Key R\&D Program of China (2020AAA0109201), NSFC grants (62022050, 62021002, 61772299), Beijing Nova Program (Z201100006820041), and MOE Innovation Plan of China.

\begin{small}
\bibliography{example_paper}
\bibliographystyle{icml2021}
\end{small}

\newpage
\appendix

\section{Dataset description and statistics}
\vspace{-5pt}

\textbf{Aircraft:} The dataset contains fine-grained classification of 10,000 aircraft pictures which belongs to 100 classes, with 100 images per class. 

\textbf{Birdsnap:} The dataset contains 49,829 images of 500 species of North American birds. 

\textbf{Caltech:} The dataset contains 9,144 pictures of objects belonging to 101 categories. There are about 40 to 800 images per category. Most categories have about 50 images.

\textbf{Cars:} The dataset contains 16,185 images of 196 classes of cars. The data is split into 8,144 training images and 8,041 testing images.

\textbf{CIFAR~10:} The dataset consists of 60,000 32x32 colorful images in 10 classes, with 6,000 images per class. There are 50,000 training images and 10,000 test images.

\textbf{CIFAR~100:} The dataset is just like the CIFAR~10, except it has 100 classes containing 600 images each.

\textbf{DTD:} The dataset contains a collection of 5,640 textural images  in the wild, annotated with a series of human-centric attributes. It has 47 classes and 120 images per class.

\textbf{Pets:} The dataset contains 7,049 images of cat and dog species which belongs to 47 classes, with around 200 images per class.

\textbf{SUN:} The dataset contains 39,700 scenery pictures with 397 classes and 100 samples per class.

For all the datasets we use, we respect the official train / val / test splits if they exist, otherwise we use $60\%$ data for training, $20\%$ data for validation (hyper-parameter tuning) and $20\%$ data for testing. 

\vspace{-15pt}

\section{Comparing LogME to re-training head}
\vspace{-5pt}

  A na\"ive way to measure the relationship between features and labels is to train a classification / regression head for the downstream task, and to use the head's performance as an assessment (sometimes it is called ``linear probing'' or ``linear protocol evaluation''). Actually we have considered this idea but find that it works not as well as expected.

  The issues of re-training head are studied by researchers in visual representation learning, too. \citet{kolesnikov_revisiting_2019} found that (1) re-training head by second-order optimization is impractical; (2) first-order optimization with gradients is sensitive to the learning rate schedule and takes a long time to converge.

  Apart from issues discussed by \citet{kolesnikov_revisiting_2019}, \citet{kornblith_better_2019} also note that hyper-parameter of logistic regression (strength of L2 regularization) should be tuned extensively, making head re-training inefficient.

\begin{figure}[htbp]
    \includegraphics[width=\columnwidth]{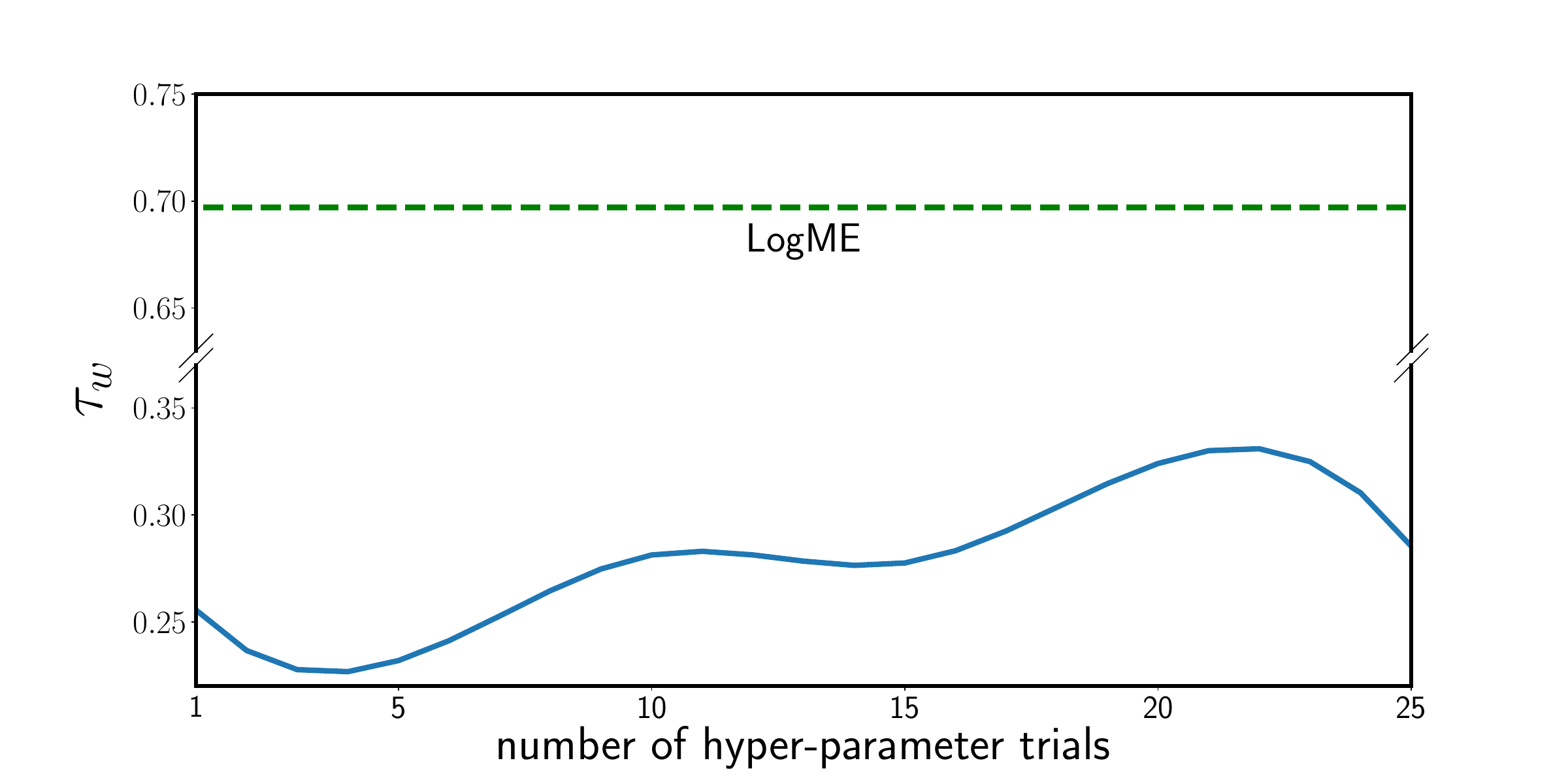}
    \vspace{-20pt}
    \caption{The correlation of re-training head \emph{w.r.t.} the number of hyper-parameter trials. It is clear that re-training head is much worse than LogME.}
    \label{fig:search_space}
  \end{figure}

  Our empirical experiments agree with the above concerns with re-training head, and also find that re-training head does not work as well as expected. In the Caltech dataset, we extract features from 10 pre-trained models, train softmax regressors with tuned hyper-parameters (the L2 regularization strength), and plot the correlation between the best head accuracy and the transfer performance \emph{w.r.t.} the number of hyper-parameter trials in Figure~\ref{fig:search_space}. The correlation of LogME is plotted as a reference. Computing LogME requires $3\times$ less time than re-training a head with one fixed hyper-parameter, and re-training head with exhaustive hyper-parameter search is still much inferior to LogME. 

  As a side issue, even if we re-train a head for the downstream task, it is unclear what quantity of the head should be used to measure pre-trained models. Since the performance of downstream tasks are evaluated by accuracy and MSE in transfer learning, it may somewhat cause over-fitting if we use the accuracy and MSE of the re-trained head. Indeed, in Figure~\ref{fig:search_space}, when the number of hyper-parameter trials increases, the correlation can even go down, showing the effect of somewhat over-fitting.

  Therefore, \emph{re-training head is neither efficient nor effective as LogME}.

  \section{Original Results in Figures}

  Original results in figures are shown in the Table~\ref{tab:figure4}, Table~\ref{tab:figure5}, and Table~\ref{tab:figure6}.

\begin{table*}[htbp]
  \centering
  \caption{Original results in Figure 4.}
  \resizebox{\textwidth}{!}{
    \begin{tabular}{clrrrrrrrrrrr}
      \toprule
    \multicolumn{2}{c}{task} & ResNet-50 & ResNet-101 & ResNet-152 & DenseNet-121 & DenseNet-169 & DenseNet-201 & Inception v1 & Inception v3 & MobileNet v2 & NASNet-A Mobile & $\tau_w$ \\
    \midrule
    \multirow{4}[0]{*}{Aircraft} & Accuracy &  86.6 & 85.6 & 85.3 & 85.4 & 84.5 & 84.6 &82.7 & 88.8 &  82.8 & 72.8 & -  \\
    \cline{3-13}
          & LEEP  &  -0.412 & -0.349 & -0.308 & -0.431 & -0.340 & -0.462 & -0.795 & -0.492 & -0.515 & -0.506 & 0.11 \\
          \cline{3-13}
          & NCE   &  -0.297 & -0.244 & -0.214 & -0.296 & -0.259 & -0.322 & -0.348 & -0.250 & -0.411 & -0.444 & 0.40\\
          \cline{3-13}
          & LogME &   0.946 & 0.948 & 0.950 & 0.938 & 0.943 & 0.942 & 0.934 & 0.953 & 0.941 & 0.948 & \textbf{0.54} \\
          \midrule
    \multirow{4}[0]{*}{Birdsnap} & Accuracy &   74.7 & 73.8 & 74.3 & 73.2 & 71.4 & 72.6 & 73.0 & 77.2 & 69.3 & 68.3 & - \\
    \cline{3-13}
          & LEEP  &   -1.647  & -1.553  & -1.481  & -1.729  & -1.756  & -1.645 & -2.483  & -1.776 & -1.951  & -1.835 &0.27 \\
          \cline{3-13}
          & NCE   &  -1.538  & -1.479  & -1.417  & -1.566  & -1.644  & -1.493  & -1.807  & -1.354 & -1.815  & -1.778  & \textbf{0.74} \\
          \cline{3-13}
          & LogME &  0.829  & 0.836  & 0.839  & 0.810  & 0.815  & 0.822  &0.806  & 0.848 & 0.808  & 0.824 & 0.67\\
          \midrule
    \multirow{4}[0]{*}{Caltech} & Accuracy &  91.8 & 93.1 & 93.2 & 91.9 & 92.5 & 93.4 & 91.7 & 94.3 & 89.1 & 91.5 & -\\
    \cline{3-13}
          & LEEP  & -2.195  & -2.067  & -1.984  & -2.159  & -2.039  & -2.122  & -2.718  & -2.286 & -2.373  & -2.263 & 0.27 \\
          \cline{3-13}
          & NCE   & -1.820  & -1.777  & -1.721  & -1.807  & -1.774  & -1.808 & -1.849  & -1.722 & -2.009  & -1.966 & 0.65 \\
          \cline{3-13}
          & LogME &   1.509  & 1.548  & 1.567  & 1.365  & 1.417  & 1.428  & 1.440  & 1.605 & 1.365  & 1.389 & \textbf{0.70}\\
          \midrule
    \multirow{4}[0]{*}{Cars} & Accuracy &   91.7  & 91.7  & 92.0  & 91.5  & 91.5  & 91.0 & 91.0  & 92.3 & 91.0  & 88.5 & - \\
    \cline{3-13}
          & LEEP  & -1.570  & -1.370  & -1.334  & -1.562  & -1.505  & -1.687  & -2.149  & -1.637  & -1.695  & -1.588 & 0.41\\
          \cline{3-13}
          & NCE   & -1.181  & -1.142  & -1.128  & -1.111  & -1.192  & -1.319  & -1.201  & -1.195 & -1.312  & -1.334 & 0.35\\
          \cline{3-13}
          & LogME &   1.253  & 1.255  & 1.260  & 1.249  & 1.252  & 1.251 & 1.246  & 1.259 & 1.250  & 1.254  & \textbf{0.65}\\
          \midrule
    \multirow{4}[0]{*}{CIFAR10} & Accuracy &  96.8  & 97.7  & 97.9  & 97.2  & 97.4  & 97.4  &96.2  & 97.5  & 95.7  & 96.8 & -\\
    \cline{3-13}
          & LEEP  &   -3.407  & -3.184  & -3.020  & -3.651  & -3.345  & -3.458 & -4.074  & -3.976 & -3.624  & -3.467 & 0.64\\
          \cline{3-13}
          & NCE   &  -3.395  & -3.232  & -3.084  & -3.541  & -3.427  & -3.467  &-3.338  & -3.625 & -3.511  & -3.436 & 0.43\\
          \cline{3-13}
          & LogME &  0.388  & 0.463  & 0.469  & 0.302  & 0.343  & 0.369  &0.293  & 0.349 & 0.291  & 0.304  & \textbf{0.78} \\
          \midrule
    \multirow{4}[0]{*}{CIFAR100} & Accuracy & 84.5  & 87.0  & 87.6  & 84.8  & 85.0  & 86.0 & 83.2  & 86.6 & 80.8  & 83.9  & -\\
    \cline{3-13}
          & LEEP  & -3.520  & -3.330  & -3.167  & -3.715  & -3.525  & -3.643  &-4.279  & -4.100 & -3.733  & -3.560 & 0.61\\
          \cline{3-13}
          & NCE   &   -3.241  & -3.112  & -2.980  & -3.304  & -3.313  & -3.323  &-3.253  & -3.447 & -3.336  & -3.254 & 0.44 \\
          \cline{3-13}
          & LogME &  1.099  & 1.130  & 1.133  & 1.029  & 1.051  & 1.061  &1.037  & 1.070 & 1.039  & 1.051 & \textbf{0.79}  \\
          \midrule
    \multirow{4}[0]{*}{DTD} & Accuracy &  75.2 & 76.2 & 75.4 & 74.9 & 74.8 & 74.5 &73.6 & 77.2 & 72.9 & 72.8 & - \\
    \cline{3-13}
          & LEEP  &    -3.663  & -3.718  & -3.653  & -3.847  & -3.646  & -3.757  & -4.124  & -4.096 & -3.805  &  -3.691  & -0.08\\
          \cline{3-13}
          & NCE   & -3.119  & -3.199  &-3.138  & -3.198  & -3.218  & -3.203  &-3.082  & -3.261 & -3.176  & -3.149 & -0.38\\
          \cline{3-13}
          & LogME &   0.761  & 0.757  & 0.766  & 0.710  & 0.730  & 0.730  &0.727  & 0.746 & 0.712  & 0.724  & \textbf{0.48}\\
          \midrule
    \multirow{4}[0]{*}{Pets} & Accuracy &   92.5  & 94.0  & 94.5  & 92.9  & 93.1  & 92.8  & 91.9  & 93.5 & 90.5  &  89.4 & -\\
    \cline{3-13}
          & LEEP  &  -1.031  & -0.915  & -0.892  & -1.100  & -1.111  & -1.108  &-1.520  & -1.129  &-1.228  & -1.150 & 0.65\\
          \cline{3-13}
          & NCE   &   -0.956  & -0.885  & -0.862  & -0.987  & -1.072  & -1.026  & -1.076  & -0.893 & -1.156  & -1.146 & \textbf{0.84}\\
          \cline{3-13}
          & LogME &    1.029  & 1.061  & 1.084  & 0.839  & 0.874  & 0.908  &0.913  & 1.191  & 0.821  & 0.833 & 0.58\\
          \midrule
    \multirow{4}[0]{*}{SUN} & Accuracy &   64.7  & 64.8  & 66.0  & 62.3  & 63.0  & 64.7  & 62.0  & 65.7 & 60.5  & 60.7 & -\\
    \cline{3-13}
          & LEEP  &  -2.611  & -2.531  & -2.513  & -2.713  & -2.570  & -2.618 & -3.153  & -2.943 & -2.764  & -2.687 & 0.58 \\
          \cline{3-13}
          & NCE   &    -2.469  & -2.455  & -2.444  & -2.500  & -2.480  & -2.465  & -2.534  & -2.529 & -2.590  & -2.586  & 0.77\\
          \cline{3-13}
          & LogME &  1.744  & 1.749  & 1.755  & 1.704  & 1.716  & 1.718  & 1.715  & 1.753 & 1.713  & 1.721 & \textbf{0.86}\\
          \bottomrule
    \end{tabular}%
  }
  \label{tab:figure4}%
\end{table*}%

\begin{table*}[htbp]
  \centering
  \caption{Original results in Figure 5.}
  \resizebox{\textwidth}{!}{
    \begin{tabular}{clrrrrrrrrrrr}
      \toprule
    \multicolumn{2}{c}{task} & ResNet-50 & ResNet-101 & ResNet-152 & DenseNet-121 & DenseNet-169 & DenseNet-201 & Inception v1 & Inception v3 & MobileNet v2 & NASNet-A Mobile & $\tau_w$ \\
    \midrule
    \multirow{2}[0]{*}{dSprites} & MSE &  0.031     &  0.028     &   0.028    &  0.039     &   0.035    &   0.036    &   0.045    &  0.044     &   0.037    &  0.035     &  - \\
    \cline{3-13}
          & LogME &  1.53     &    1.64   &   1.63    &   1.35    &    1.25   &   1.34    &  1.18     &  1.22     &   1.18    &   1.39    & 0.84 \\
          \bottomrule
    \end{tabular}%
  }
  \label{tab:figure5}%
\end{table*}%

\begin{table*}[htbp]
  \centering
  \caption{Original results in Figure 6.}
  \resizebox{\textwidth}{!}{
    \begin{tabular}{clrrrrrrrrr}
      \toprule
    \multicolumn{2}{c}{task} & RoBERTa & RoBERTa-D & uncased BERT-D & cased BERT-D & ALBERT-v1 & ALBERT-v2 & ELECTRA-base & ELECTRA-small & $\tau_w$ \\
    \midrule
    \multirow{2}[0]{*}{MNLI} & Accuracy &  87.6    & 84.0  & 82.2 & 81.5 & 81.6& 84.6& 79.7& 85.8& -     \\
    \cline{3-11}
          & LogME &  -0.568   &   -0.599    &   -0.603    &  -0.612&  -0.614&-0.594&  -0.666&-0.621 & 0.66 \\
          \midrule
    \multirow{2}[0]{*}{QQP} & Accuracy & 91.9 & 89.4 & 88.5 & 87.8    &  -     &  -     &   -    &    - & -   \\
    \cline{3-11}
          & LogME &  91.9 & 89.4 & 88.5 & 87.8            &   -    & -      &    -   &      - &  0.73\\
          \midrule
    \multirow{2}[0]{*}{QNLI} & Accuracy &  92.8 & 90.8 & 89.2 & 88.2   &  -     &    -   &     -  &    -   &    -   \\
    \cline{3-11}
          & LogME &  -0.565    &  -0.603    &   -0.613    &   -0.618            & -      &      - &   -    &     -  & 1.00 \\
          \midrule
    \multirow{2}[0]{*}{SST-2} & Accuracy &    94.8 & 92.5 & 91.3 & 90.4 & 90.3 & 92.9   & - & - & -    \\
    \cline{3-11}
          & LogME &   -0.312 & -0.330   &   -0.331    &    -0.353& -0.525&-0.447       &    -   &   -    & 0.68 \\
          \midrule
    \multirow{2}[0]{*}{CoLA} & Accuracy &  63.6 & 59.3 & 51.3 & 47.2  & - &- &- &- & -  \\
    \cline{3-11}
          & LogME &   -0.499   &    -0.536   &   -0.568    &  -0.572            &  -     &   -    &   -    &     -  & 1.00 \\
          \midrule
    \multirow{2}[0]{*}{MRPC} & Accuracy &   90.2& 86.6 & 87.5 & 85.6 &- &- &- &- &-    \\
    \cline{3-11}
          & LogME &  -0.573  &   -0.586    & -0.605      &    -0.604           &   -    &     -  &   -    &    -   &  0.53\\
          \midrule
    \multirow{2}[0]{*}{RTE} & Accuracy & 78.7 & 67.9& 59.9    & 60.6      &    -   &  -     &  -     & -      &   -    \\
    \cline{3-11}
          & LogME &    -0.709 &  -0.723     &    -0.725   &    -0.725           &   -    &  -     &   -    &    -   &  1.00\\
          \bottomrule
    \end{tabular}%
  }
  \label{tab:figure6}%
\end{table*}%

\end{document}